\documentclass[letterpaper]{article} 
\usepackage{aaai2027}  
\usepackage[hyphens]{url}  
\usepackage{graphicx} 
\usepackage{natbib}  
\usepackage{caption} 
\usepackage{algorithm}
\usepackage{algorithmic}

\usepackage{newfloat}
\usepackage{listings}
\DeclareCaptionStyle{ruled}{labelfont=normalfont,labelsep=colon,strut=off} 
\floatstyle{ruled}
\newfloat{listing}{tb}{lst}{}
\floatname{listing}{Listing}

\usepackage{booktabs}
\usepackage{amsfonts}
\usepackage{amsmath} 
\usepackage{amssymb}
\nocopyright
\title{NeuroPB: Scaling Neural Decoding with Pretrained Behavioral Representations}
\author{
    Luyao Jin\textsuperscript{\rm 1},
    Yonghao Song\textsuperscript{\rm 2},
    Huan Zhao\textsuperscript{\rm 1},
    Vincent C. K. Cheung\textsuperscript{\rm 3},
    Wei-Hsin Liao\textsuperscript{\rm 1}\corresponding
}
\affiliations{
    \textsuperscript{\rm 1}Department of Mechanical and Automation Engineering, The Chinese University of Hong Kong\\
    \textsuperscript{\rm 2}Department of Biomedical Engineering, School of Medicine, Tsinghua University\\
    \textsuperscript{\rm 3}School of Biomedical Sciences, The Chinese University of Hong Kong\\

    lyjin@link.cuhk.edu.hk,
    whliao@cuhk.edu.hk
}

\begin{document}

\maketitle

\begin{abstract}
Decoding continuous motor trajectories from neural activity is essential for developing practical brain–computer interfaces (BCIs). However, current neural decoders are constrained by the limited scale and heterogeneity of neural recordings. In contrast, behavioral data can be collected more readily and at substantially larger scale from humans, animals, simulations, and robotic systems. Here, we introduce NeuroPB, a framework that scales neural decoding by transferring knowledge from pretrained behavioral representations. NeuroPB first pretrains a motor encoder on large-scale motor behavior data and then aligns neural activity with the resulting behavioral representation space using a limited set of paired neural–behavioral recordings. A neural encoder and lightweight motor decoder are subsequently optimized to reconstruct continuous movement from the aligned neural representations. Across multiple macaque motor datasets, behavioral pretraining improves trajectory decoding, including an 11\%  $R^2$ increase on center-out and 8\% on random-target compared with training the motor encoder from scratch. Notably, pretraining on robotic trajectories achieves performance comparable to pretraining on macaque trajectories, demonstrating that transferable kinematic structure is shared across biological and artificial models. Moreover, decoding performance improves as the scale and diversity of robotic pretraining data increase, when the amount of neural data is fixed. Pretraining also enhances generalization across recording sessions, subjects, and motor tasks, with only 10\% calibration needed to match training from scratch. Overall, these results establish behavioral pretraining as a scalable source for neural decoding and provide a promising route toward high-performance and calibration-efficient BCIs under limited neural data.
\end{abstract}


\section{Introduction}
Brain-computer interfaces (BCIs) provide a direct pathway for translating neural activity into executable commands, enabling users to interact with prosthetic limbs, robotic systems, computers, and other external devices \cite{bci}. Among the capabilities required by practical BCIs, decoding continuous motor behavior is particularly important because natural interaction depends not only on identifying a discrete intention, but also on reconstructing how a movement evolves over time \cite{lee2025brain, willsey2025high}. Despite substantial progress, accurately decoding continuous behavioral trajectories from neural activity remains challenging. Neural activity is heterogeneous across recording sessions, individuals, and motor tasks, while collecting sufficiently large and well-annotated neural datasets is expensive and technically demanding.

\begin{figure}[t]
\centering
\includegraphics[width=0.9\columnwidth]{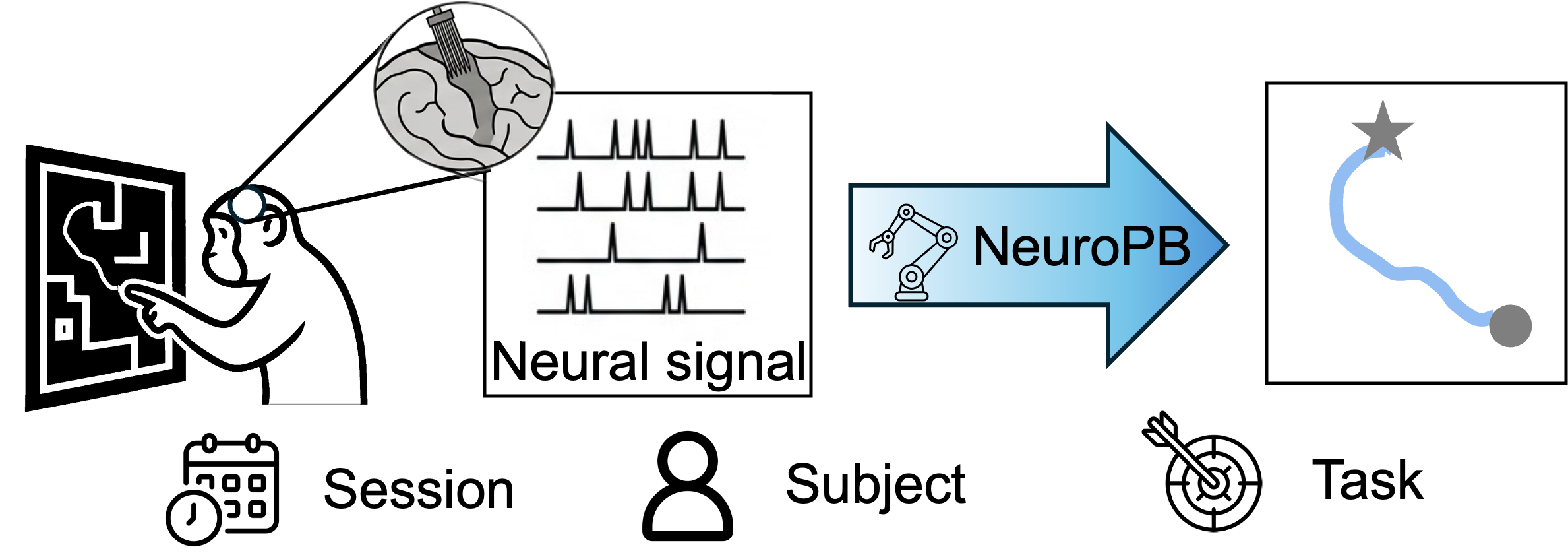}
\caption{Conceptual overview of NeuroPB. NeuroPB aligns neural activity with a behavioral representation pretrained on scalable macaque and robotic trajectories, enabling continuous trajectory decoding with generalization across recording sessions, subjects, and motor tasks.}
\label{fig1}
\end{figure}

Existing neural decoding methods learn direct mappings from neural activity to movement trajectories using end-to-end models \cite{LFADS, 2020machine, dad}. Although these methods achieve good performance within the training distribution, their performance often degrades when applied to unseen recording sessions, subjects, or motor tasks. To improve generalization, recent studies have explored large-scale neural pretraining \cite{ned, poyo+} and self-supervised learning \cite{cebra}. These approaches aim to learn general neural representations before adapting the model to a downstream decoding task. However, their scalability remains constrained by the limited availability and pronounced heterogeneity of neural recordings across subjects, sessions, brain regions, and acquisition devices. Consequently, increasing the scale of neural pretraining alone may not provide a readily scalable solution. This limitation motivates us to hypothesize: rather than relying exclusively on scarce and heterogeneous neural recordings, can prior knowledge about behavior itself be used to guide the learning of more transferable neural representations?

In this work, we approach neural decoding from the behavioral side. Recent advances in embodied AI have promoted large, diverse, and standardized manipulation trajectories \cite{libero, bridgedata, oxe}. Compared with neural recordings, robotic behavior can be collected at substantially larger scales \cite{ndt2, oxe}. Biological movements, in contrast, naturally contain trial-to-trial variability \cite{fishbach2007deciding}. Despite their differences in embodiment, biological and artificial systems share motor control objectives and can exhibit reusable behavioral structure \cite{merel2019hierarchical}. We therefore hypothesize that representations learned from large-scale behavioral trajectories can provide transferable motor priors for neural decoding.

To investigate this hypothesis, we propose \textbf{NeuroPB}, a \textbf{Neuro}n decoding framework with \textbf{P}retrained \textbf{B}ehavioral representation that bridges neural activity with representations extracted by a pretrained behavior model. NeuroPB uses a pretrained motor encoder to define a structured behavioral representation space. During training, a neural encoder maps neural activity into this space, allowing pretrained behavioral representations to serve as anchors that guide the neural encoder toward capturing motor-relevant information. The resulting neural representation is then passed to a motor decoder to reconstruct the corresponding continuous movement. At inference time, the pretrained motor encoder is no longer required: the neural encoder directly extracts a motor-informed representation from the neural signal, and the motor decoder generates the predicted trajectory. In this way, NeuroPB transfers behavioral knowledge to the neural decoding model without requiring behavioral inputs during deployment.


We evaluate NeuroPB on multiple macaque neural datasets covering different recording sessions, subjects, and motor tasks. Our experiments show that pretrained behavioral representations consistently facilitate neural trajectory decoding and that scaling trajectory pretraining leads to stronger overall performance. Notably, trajectory encoders pretrained on large-scale robotic data provide effective priors for decoding biological movements, supporting the existence of transferable behavioral structure across embodiments. NeuroPB also exhibits improved generalization to unseen sessions, subjects, and tasks, while requiring fewer paired neural and trajectory samples during calibration. The main contributions are summarized as follows:

\begin{itemize}
    \item We propose the \textbf{NeuroPB}, a neural decoding framework that uses pretrained behavioral representations as anchors to learn motor-informed neural representations.
    \item We demonstrate that behavioral pretraining consistently outperforms training the framework from scratch, establishing the importance of behavioral pretraining for neural trajectory decoding.
    \item We investigate the scaling of the pretrained motor encoder and show that larger-scale and diverse behavioral datasets yield better decoding performance.
    \item We evaluate NeuroPB on macaque datasets, demonstrating improved calibration efficiency and robust generalization across recording sessions, subjects, and motor tasks.
\end{itemize}

\section{Related Work}
\subsection{Motor Neural Decoding}
Motor neural decoding seeks to infer behavioral variables from neural population activity. Early approaches map neural activity to movement trajectories, such as Wiener filters \cite{Wiener}, Kalman filters \cite{2020machine}, statistical models \cite{LFADS}, recurrent networks \cite{ndt}, sequential autoencoders \cite{autolfads}, and other deep architectures \cite{OrthoSchema}. Although these methods can achieve high decoding accuracy within the recording sessions, they often generalize poorly to unseen sessions, subjects, or motor tasks. To improve generalization, recent studies have pretrained neural encoders on recordings collected from multiple sessions, subjects, and tasks. NEDS learns generalized latent population dynamics across datasets \cite{ned}, CEBRA constructs behaviorally structured neural representations using contrastive learning \cite{cebra}, and large-scale frameworks such as POYO+ and NDT3 integrate heterogeneous neural recordings through neural pretraining \cite{poyo+, ndt3}. These approaches demonstrate that increasing the scale of neural pretraining can improve adaptation to new recording sessions. Nevertheless, their scalability remains dependent on the availability of large, high-quality neural datasets, which are costly to collect and heterogeneous across subjects, sessions, brain regions, and acquisition devices. Our work therefore explores a complementary direction: learning transferable behavioral representations from large-scale behavioral trajectories and using them to guide neural representation learning with limited paired neural–behavioral data.

\subsection{Contrastive Learning in Neural Decoding}
Contrastive learning provides an effective mechanism for neural decoding by bringing matched neural–target representations closer while separating mismatched samples, thereby emphasizing information shared across modalities. Inspired by the CLIP framework for large-scale vision–language alignment~\cite{clip}, this strategy has been increasingly adopted to connect neural signals with behaviorally meaningful representation spaces. CEBRA uses contrastive sampling conditioned on behavior or time to learn consistent neural embeddings across sessions, animals, and recording modalities~\cite{cebra}. Contrastive alignment is employed between neural and speech representations for speech decoding ~\cite{speech_decoding, bit}. Similar ideas have been explored in visual decoding, where aligned EEG/fMRI responses with image embeddings are used for object recognition~\cite{nice, zhang2026neurobridge, umbrae}. Unlike prior work that relies primarily on paired neural–target data, our framework also leverages independently collected, large-scale robotic trajectories to pretrain a scalable behavioral representation, which is then aligned with neural activity for transferable motor decoding. This design enables neural decoding to benefit from scalable behavioral data beyond biological recordings.

\begin{figure*}[t]
\centering
\includegraphics[width=0.8\textwidth]{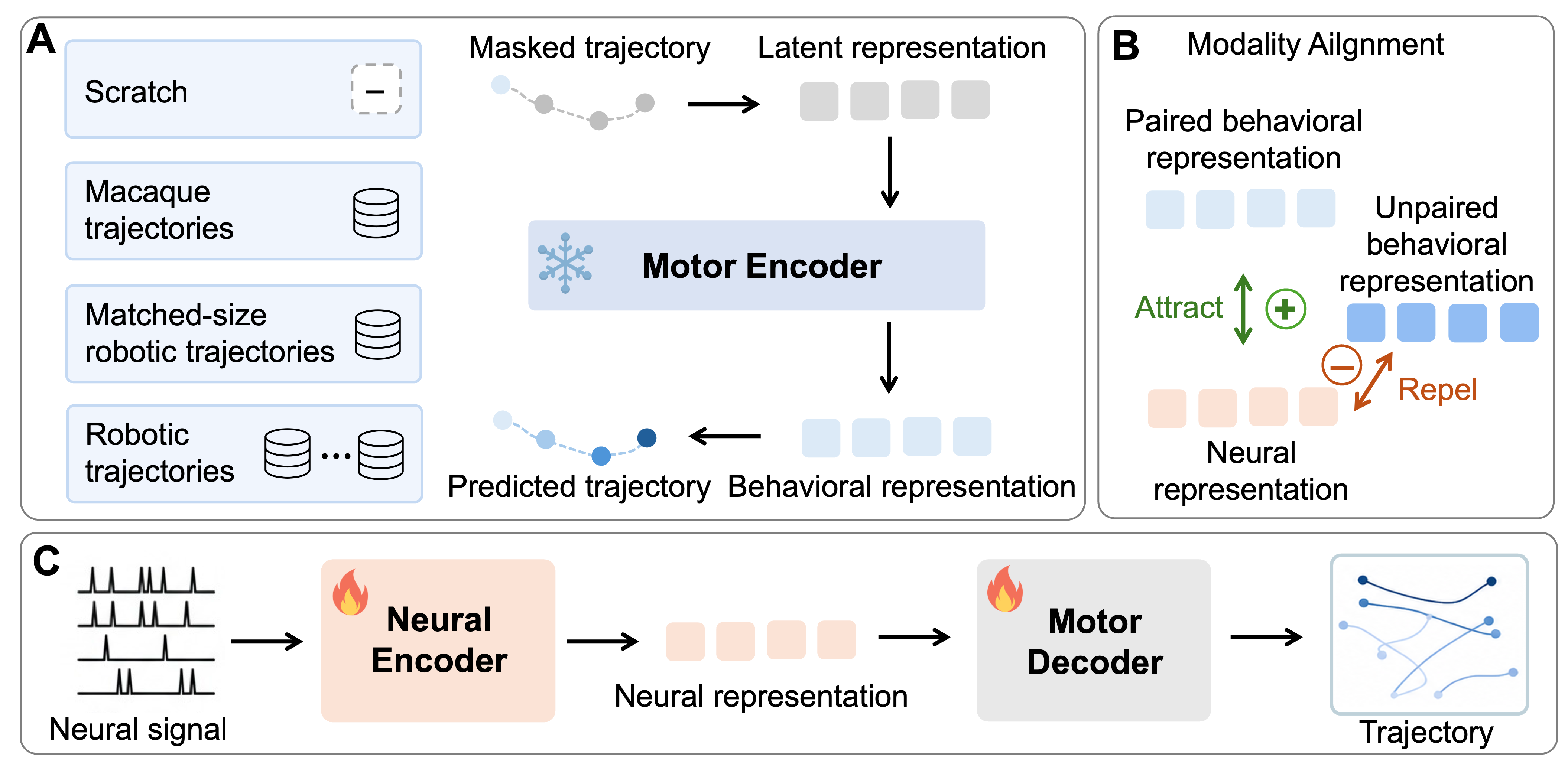} 
\caption{(A) A motor encoder is pretrained on masked trajectory reconstruction using macaque or robotic trajectories and then frozen to produce behavioral representations. (B) Contrastive learning aligns paired neural and behavioral representations while separating unpaired samples. (C) During inference, the neural encoder maps spike activity into the neural representation, from which a motor decoder reconstructs movement trajectories.}
\label{fig2}
\end{figure*}

\section{Method}
Our framework learns transferable neural representations by using a pretrained behavioral representation space as behavioral supervision. As illustrated in Fig. \ref{fig2}, the method consists of three stages. First, the motor encoder is pretrained through masked trajectory reconstruction using macaque trajectories, scale-matched robotic trajectories, or large-scale robotic trajectories, while a randomly initialized encoder serves as the scratch condition. The resulting motor encoder is then kept frozen. Second, the neural encoder is optimized with a contrastive objective that aligns matched neural and behavioral representations while separating mismatched pairs. Finally, a lightweight motor decoder reconstructs the continuous movement trajectory from the aligned neural representation. Detailed network architectures are provided in the Appendix.

\subsection{Motor Encoder}
We construct the motor encoder based on the Transformer structure, which learns structured movement representations through masked trajectory reconstruction \cite{atm}. Given a trajectory segment
\begin{equation}
    \mathbf{P}
    =
    [\mathbf{p}_{1},\mathbf{p}_{2},\ldots,\mathbf{p}_{T}],
    \qquad
    \mathbf{p}_{t}\in\mathbb{R}^{d_p},
\end{equation}
where $\mathbf{p}_{t}$ denotes the movement position at time step $t$ and
$d_p=2$ for the $(x,y)$, a subset of
time steps $\mathcal{M}\subseteq\{1,\ldots,T\}$ is randomly masked. The input
token at each time step is defined as
\begin{equation}
    \widetilde{\mathbf{e}}_{t}
    =
    \begin{cases}
        E_{\mathrm{pos}}(\mathbf{p}_{t})
        + E_{\mathrm{time}}(t),
        & t\notin\mathcal{M}, \\[4pt]
        \mathbf{e}_{\mathrm{mask}}
        + E_{\mathrm{time}}(t),
        & t\in\mathcal{M},
    \end{cases}
\end{equation}
where $E_{\mathrm{pos}}(\cdot)$ projects trajectory coordinates into the model
dimension, $E_{\mathrm{time}}(\cdot)$ encodes temporal order, and
$\mathbf{e}_{\mathrm{mask}}$ is a learnable mask embedding.
The resulting token sequence is processed by the motor encoder
$f_{\theta_m}$:
\begin{equation}
    \mathbf{H}^{m}
    =
    f_{\theta_m}
    \left(
        \widetilde{\mathbf{e}}_{1},
        \widetilde{\mathbf{e}}_{2},
        \ldots,
        \widetilde{\mathbf{e}}_{T}
    \right),
\end{equation}
where
\begin{equation}
    \mathbf{H}^{m}
    =
    [\mathbf{h}^{m}_{1},\mathbf{h}^{m}_{2},\ldots,\mathbf{h}^{m}_{T}]
    \in\mathbb{R}^{T\times D}
\end{equation}
denotes the contextualized trajectory tokens. A lightweight prediction head
$g_{\phi}$ reconstructs the masked trajectory coordinates:
\begin{equation}
    \widehat{\mathbf{p}}_{t}
    =
    g_{\phi}\left(\mathbf{h}^{m}_{t}\right),
    \qquad t\in\mathcal{M}.
\end{equation}
The motor encoder is pretrained using the masked trajectory reconstruction
objective
\begin{equation}
    \mathcal{L}_{\mathrm{traj}}
    =
    \frac{1}{|\mathcal{M}|}
    \sum_{t\in\mathcal{M}}
    \left\|
        \widehat{\mathbf{p}}_{t}
        -
        \mathbf{p}_{t}
    \right\|_{2}^{2}.
\end{equation}

The behavioral representation of the trajectory is obtained by aggregating
the contextualized trajectory tokens:
\begin{equation}
    \mathbf{z}^{m}
    =
    \operatorname{Pool}\left(\mathbf{H}^{m}\right),
    \qquad
    \overline{\mathbf{z}}^{m}
    =
    \frac{\mathbf{z}^{m}}
    {\left\|\mathbf{z}^{m}\right\|_{2}},
\end{equation}
where $\operatorname{Pool}(\cdot)$ denotes token aggregation and
$\overline{\mathbf{z}}^{m}$ is the normalized behavioral representation used as the target for neural--behavioral alignment. 

To disentangle the effects of behavioral source and pretraining scale, we evaluated four motor encoder initialization strategies: training from scratch (Scratch), pretraining on macaque trajectories (Macaque-PT), pretraining on a scale-matched robotic trajectory subset (Robot-Matched-PT), and pretraining on the large-scale LIBERO-100 robotic trajectory dataset (Robot-Large-PT). In Scratch, the motor encoder was randomly initialized and jointly optimized during neural alignment. Macaque-PT, Robot-Matched-PT, and Robot-Large-PT mean pretrained with macaque trajectories, robotic trajectories matched in size with macaque trajectories, and larger-scale robotic trajectories, respectively. Then, keep frozen during neural alignment. Comparing Macaque-PT with Robot-Matched-PT isolates the effect of behavioral source under a matched data scale, whereas comparing Robot-Matched-PT with Robot-Large-PT evaluates the effect of increasing robotic pretraining scale.

\subsection{Neural Encoder}
We construct the neural encoder following the Perceiver-based POYO architecture~\cite{poyo, perceiverio}, which represents neural population activity as a sequence of timestamped spike events and compresses the variable-length sequence into a fixed number of latent tokens. For a neural window of duration $T$, let
\begin{equation}
    \mathcal{S}
    =
    \left\{
        (u_i,t_i)
    \right\}_{i=1}^{M}
\end{equation}
denote the $M$ observed spike events, where $u_i$ is the identity of the neural unit that generated the $i$-th spike and $t_i\in[0,T]$ is its timestamp. Each spike event is represented using a learnable unit embedding:
\begin{equation}
    \mathbf{x}_{i}
    =
    E_{\mathrm{unit}}(u_i),
    \qquad
    \mathbf{X}
    =
    [\mathbf{x}_{1},\mathbf{x}_{2},\ldots,\mathbf{x}_{M}]
    \in\mathbb{R}^{M\times D}.
\end{equation}
Temporal information is incorporated through rotary positional embeddings \cite{roformer}, allowing the attention operation to depend on the relative timing between spike events and latent tokens. For adapt to different numbers of spikes varies across windows, sessions, and neural populations, we introduce $N$ latent tokens
\begin{equation}
    \mathbf{Z}_{0}
    =
    [\mathbf{z}_{0,1},\mathbf{z}_{0,2},\ldots,\mathbf{z}_{0,N}]
    \in\mathbb{R}^{N\times D},
    \qquad N\ll M,
\end{equation}
whose timestamps are uniformly distributed over the neural context window. The latent tokens first aggregate information from the spike-event sequence through cross-attention:
\begin{equation}
    \mathbf{Z}_{1}
    =
    \operatorname{CrossAttn}
    \left(
        \mathbf{Z}_{0},
        \mathbf{X}
    \right).
\end{equation}
More specifically, the cross-attention operation is given by
\begin{equation}
\begin{aligned}
    \operatorname{softmax}
    \left(
        \frac{
            \operatorname{RoPE}
            \left(
                \mathbf{Z}_{0}\mathbf{W}_{q},
                \boldsymbol{\tau}_{z}
            \right)
            \operatorname{RoPE}
            \left(
                \mathbf{X}\mathbf{W}_{k},
                \boldsymbol{\tau}_{x}
            \right)^{\top}
        }
        {\sqrt{d_k}}
    \right)
    \mathbf{X}\mathbf{W}_{v},
\end{aligned}
\end{equation}
where $\boldsymbol{\tau}_{x}$ and $\boldsymbol{\tau}_{z}$ denote the timestamps assigned to the spike tokens and latent tokens, respectively, and $\mathbf{W}_{q}$, $\mathbf{W}_{k}$, and $\mathbf{W}_{v}$ are learnable projection matrices. The compressed latent sequence is subsequently processed by $L-1$ Transformer self-attention blocks,
The output of the final self-attention block as the neural representation:
\begin{equation}
    \mathbf{Z}^{n}
    =
    \mathbf{Z}_{L}
    =
    [\mathbf{z}_{L,1},\mathbf{z}_{L,2},\ldots,\mathbf{z}_{L,N}]
    \in\mathbb{R}^{N\times D}.
\end{equation}
Thus, $\mathbf{Z}_{L}$ summarizes the temporal and population-level structure of the neural activity while maintaining a fixed dimensionality across recording sessions with different numbers and identities of neural units. For modality alignment, the latent tokens are aggregated and transformed by a learnable neural projection head $g_{n}$:
\begin{equation}
    \mathbf{z}^{n}
    =
    \operatorname{Pool}\left(\mathbf{Z}^{n}\right),
    \qquad
    \widetilde{\mathbf{z}}^{n}
    =
    g_{n}\left(\mathbf{z}^{n}\right),
\end{equation}
followed by $\ell_2$ normalization:
\begin{equation}
    \overline{\mathbf{z}}^{n}
    =
    \frac{\widetilde{\mathbf{z}}^{n}}
    {\left\|\widetilde{\mathbf{z}}^{n}\right\|_{2}}.
\end{equation}
The resulting representation $\overline{\mathbf{z}}^{n}$ is aligned with the corresponding behavioral representation produced by the frozen motor encoder.
\subsection{Modality Alignment}
After obtaining neural and behavioral representations, we align the two modalities using a CLIP-style symmetric contrastive objective~\cite{clip}. Given a mini-batch of $B$ temporally paired neural and behavioral samples,
\begin{equation}
    \left\{
        \left(
            \mathcal{S}_{i},
            \mathbf{P}_{i}
        \right)
    \right\}_{i=1}^{B},
\end{equation}
the neural encoder and frozen motor encoder get normalized behavioral representation and neural representation, 
\begin{equation}
    \left\{
        \left(
            \overline{\mathbf{z}}^{m}_{i},
            \overline{\mathbf{z}}^{n}_{i}
        \right)
    \right\}_{i=1}^{B},
\end{equation}
We compute the cross-modal similarity between the $i$-th neural representation and the $j$-th behavioral representation using scaled cosine similarity:
\begin{equation}
    s_{ij}
    =
    \frac{
        \left(
            \overline{\mathbf{z}}^{n}_{i}
        \right)^{\top}
        \overline{\mathbf{z}}^{m}_{j}
    }{\tau},
\end{equation}
where $\tau$ is a temperature parameter controlling the concentration of the similarity distribution. For each neural representation, its temporally paired behavioral representation is treated as the positive sample, whereas the remaining behavioral representations in the mini-batch are treated as negative samples. The neural-to-motor contrastive loss is defined as
\begin{equation}
    \mathcal{L}_{n\rightarrow m}
    =
    -\frac{1}{B}
    \sum_{i=1}^{B}
    \log
    \frac{
        \exp(s_{ii})
    }{
        \sum_{j=1}^{B}
        \exp(s_{ij})
    }.
\end{equation}
Similarly, by treating each behavioral representation as a query and the neural representations as candidates, the motor-to-neural contrastive loss is
\begin{equation}
    \mathcal{L}_{m\rightarrow n}
    =
    -\frac{1}{B}
    \sum_{i=1}^{B}
    \log
    \frac{
        \exp(s_{ii})
    }{
        \sum_{j=1}^{B}
        \exp(s_{ji})
    }.
\end{equation}
The final modality-alignment objective is the symmetric average of the two directions:
\begin{equation}
    \mathcal{L}_{\mathrm{align}}
    =
    \frac{1}{2}
    \left(
        \mathcal{L}_{n\rightarrow m}
        +
        \mathcal{L}_{m\rightarrow n}
    \right).
\end{equation}

During modality alignment, the pretrained motor encoder remains frozen,
whereas the neural encoder and the neural projection head $g_{n}$ are
optimized. The objective increases the similarity between temporally paired neural and behavioral representations while reducing the similarity between mismatched pairs. Consequently, the neural encoder is encouraged to preserve behaviorally relevant information shared with the pretrained behavioral representation space while suppressing recording-specific variability across units and sessions.
\subsection{Motor Decoder}
The motor decoder reconstructs the continuous movement trajectory from the aligned neural representation. Given the pooled neural representation $\overline{\mathbf{z}}^{n}_{i}$, we employ a lightweight multilayer perceptron $d_{\psi}$ to predict the trajectory: 
\begin{equation}
    \widehat{\mathbf{P}}_{i}
    =
    d_{\psi}\left(\overline{\mathbf{z}}^{n}_{i}\right)
\end{equation}
It was trained jointly with the neural encoder; the total objective was
\begin{equation}\mathcal{L}=\mathcal{L}_{\mathrm{align}}+\mathcal{L}_{\mathrm{motor}}
\end{equation}
where \(\mathcal{L}_{\mathrm{motor}}\) was the mean-squared error between the predicted and ground truth positions \(\mathcal{L}_{\mathrm{motor}}=\mathrm{MSE}( \widehat{\mathbf{P}},\mathbf{P})\).

\begin{table}[t]
\centering
\begin{tabular}{l l l}
  \toprule Method & CO ($R^2$ ) & RT ($R^2$ ) \\ \midrule Wiener Filter & 0.5010$\pm$0.0678 & 0.4240$\pm$0.0838 \\ Smoothing & 0.6457$\pm$0.0606 & 0.5588$\pm$0.0710 \\DenseNN & 0.6261$\pm$0.0637 & 0.5681$\pm$0.0599 \\RNN & 0.7028$\pm$0.0552 & 0.6439$\pm$0.0576 \\ LFADS  & 0.7813$\pm$0.1020 & 0.6254$\pm$0.0765 \\
  NEDS & 0.7635$\pm$0.0758 & 0.6121$\pm$0.0918\\
  NDT2 & 0.9258$\pm$0.0031 & 0.6323$\pm$0.1339\\
  POYO & 0.9427$\pm$0.0019 & 0.7156$\pm$0.0966 \\ NeuroPB & \textbf{0.9478$\pm$0.0120} & \textbf{0.8475$\pm$0.0171} \\
  \bottomrule 
\end{tabular}
\caption{Comparison with baselines on single-session.}
\label{table1}
\end{table}

\section{Experiments and Results}
\subsection{Datasets}
We evaluated our method on two datasets, a large-scale collection of electrophysiological and behavioral recordings from four rhesus macaques performing two-dimensional reaching tasks \cite{dataset}. Neural activity was recorded using chronically implanted multielectrode arrays in the primary motor cortex (M1) and dorsal premotor cortex (PMd), with some sessions containing simultaneous recordings from both regions and up to 192 electrodes. It contains two behavioral paradigms: center-out reaching (CO), in which the macaque moved a cursor from a central location toward one of eight peripheral targets arranged radially, and random-target reaching (RT), in which the macaque continuously acquired targets presented at varying spatial locations. The dataset provides spike events together with synchronized cursor position and task-related metadata. The recordings span different collection sessions, subjects, and task structures, making them suitable for evaluating both within-domain decoding and generalization across sessions, subjects, and motor tasks.
\subsection{Implement Details}
Our method was implemented in PyTorch on an RTX 4090 GPU. We used the SparseLamb optimizer with a weight decay of \(10^{-4}\). The maximum learning rate was \(3.125\times10^{-4}\times128=0.04\), where 128 is the batch size. The learning rate followed a one-cycle schedule with cosine annealing. After training, the checkpoint with the highest average validation metric was evaluated once on the held-out test set. Additional experimental details are provided in the Appendix. In all experiments, the training, validation, and test sets were defined by the predefined split-specific sampling intervals in the dataset configuration, with around 10\% holdout for validation and 20\% holdout for testing. The 1s window is used to sample from the training intervals. Validation and test windows used 50\% overlap within the same split. For cross experiments, the split was performed at the corresponding session level, such that all data from a given session were assigned to only one partition.

\subsection{Comparison with Baselines}
We compared NeuroPB with several classical and recent models on the center-out (CO) and random-target (RT) tasks. Table \ref{table1} shows the performance from Wiener Filter \cite{Wiener}, Smoothing, DenseNN, RNN \cite{2020machine}, and LFADS \cite{LFADS}, which were reproduced using the same protocol. NEDS \cite{ned}, NDT2 \cite{ndt2}, and POYO \cite{poyo} results were obtained from other publications \cite{possm, mojo}. NeuroPB achieved the state-of-the-art decoding performance on both datasets, reaching 94.78\%±1.20\% $R^2$ on CO and 84.75\%±1.71\% $R^2$ on RT. The results indicate that NeuroPB remains effective for tasks with different movement structures. While several baselines performed reasonably well on the more structured CO task, their performance generally decreased on RT, which contains more variable target locations and trajectories. NeuroPB maintained strong performance on both tasks and showed low variability across evaluation sessions. These findings demonstrate that aligning neural activity with a pretrained behavioral representation provides a more accurate and robust basis for continuous trajectory decoding than direct end-to-end decoding or neural-only representation learning.

\begin{figure}[t]
\centering
\includegraphics[width=1.0\columnwidth]{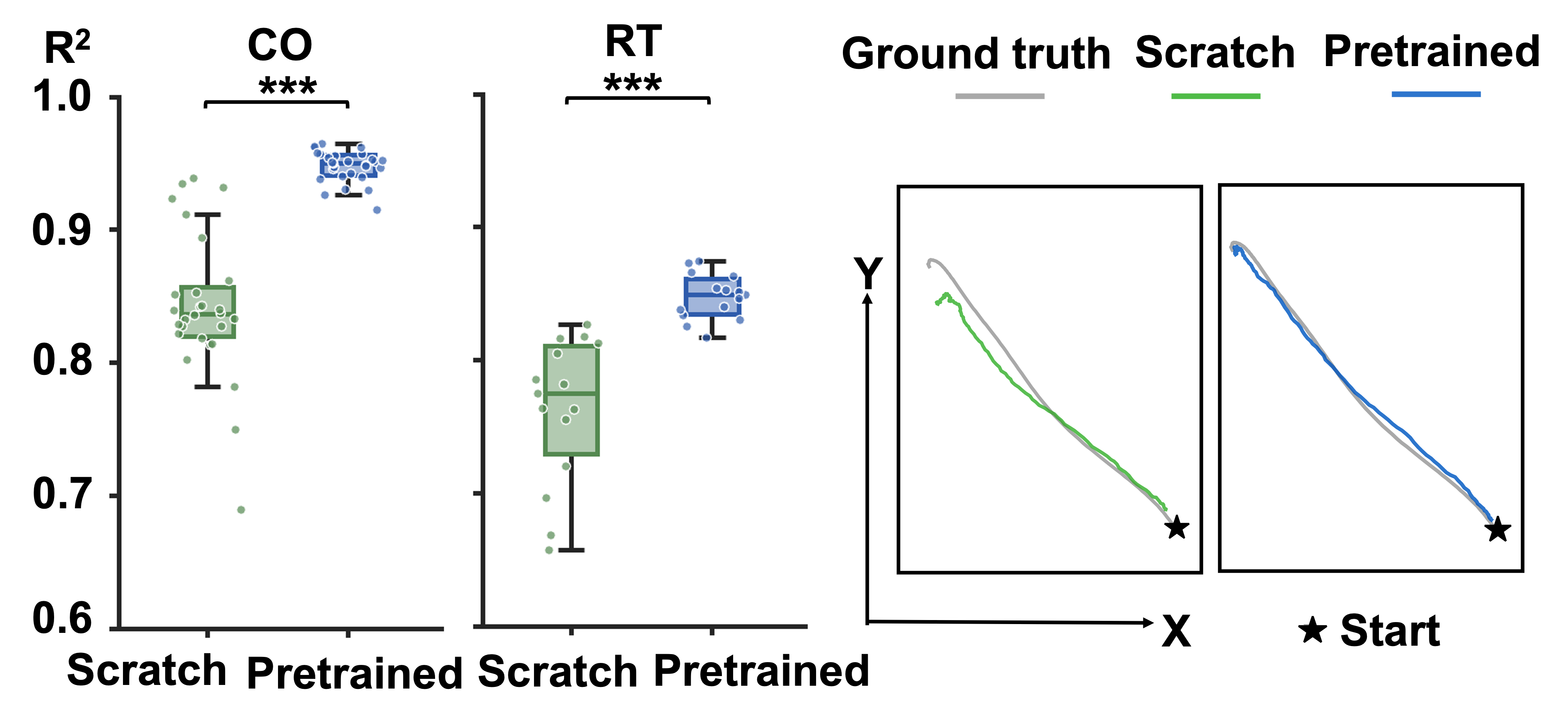}
\caption{Robot-Large-PT significantly outperforms Scratch on CO and RT and produces trajectories that more closely match the ground truth. *** denotes $p<0.001$.}
\label{fig3}
\end{figure}

\subsection{Effect of Pretrained Motor Encoder}
In the overall evaluation, NeuroPB uses the Robot-Large-PT motor encoder. To isolate the contribution of behavioral pretraining, we compared this initialization with Scratch, which means the motor encoder was initialized randomly and jointly optimized with neural alignment. As shown in the left panel of Fig. \ref{fig3}, behavioral pretraining substantially improved decoding performance on both datasets. Statistical significance was assessed using a two-sided paired t-test. On CO, Robot-Large-PT achieved an $R^2$ of 94.78\% compared with 84.19\% for scratch. On RT, Robot-Large-PT achieved 84.75\%, outperforming scratch at 76.29\%. In addition to improving the average decoding accuracy, pretraining markedly reduced performance variability across evaluations, suggesting that the pretrained behavioral representation provides a more stable target space for neural–behavioral alignment. 

The representative trajectory examples further illustrate this improvement in the right panel of Fig. \ref{fig3}. Predictions produced with the scratch motor encoder exhibit larger deviations from the ground-truth movement paths, whereas those produced with Robot-Large-PT more closely follow the overall direction and geometry of the movements toward the target. Together, these results demonstrate that the performance of NeuroPB does not arise solely from its alignment architecture: behavioral pretraining is a critical component for learning accurate and robust neural trajectory decoders.

\begin{figure}[t]
\centering
\includegraphics[width=1.0\columnwidth]{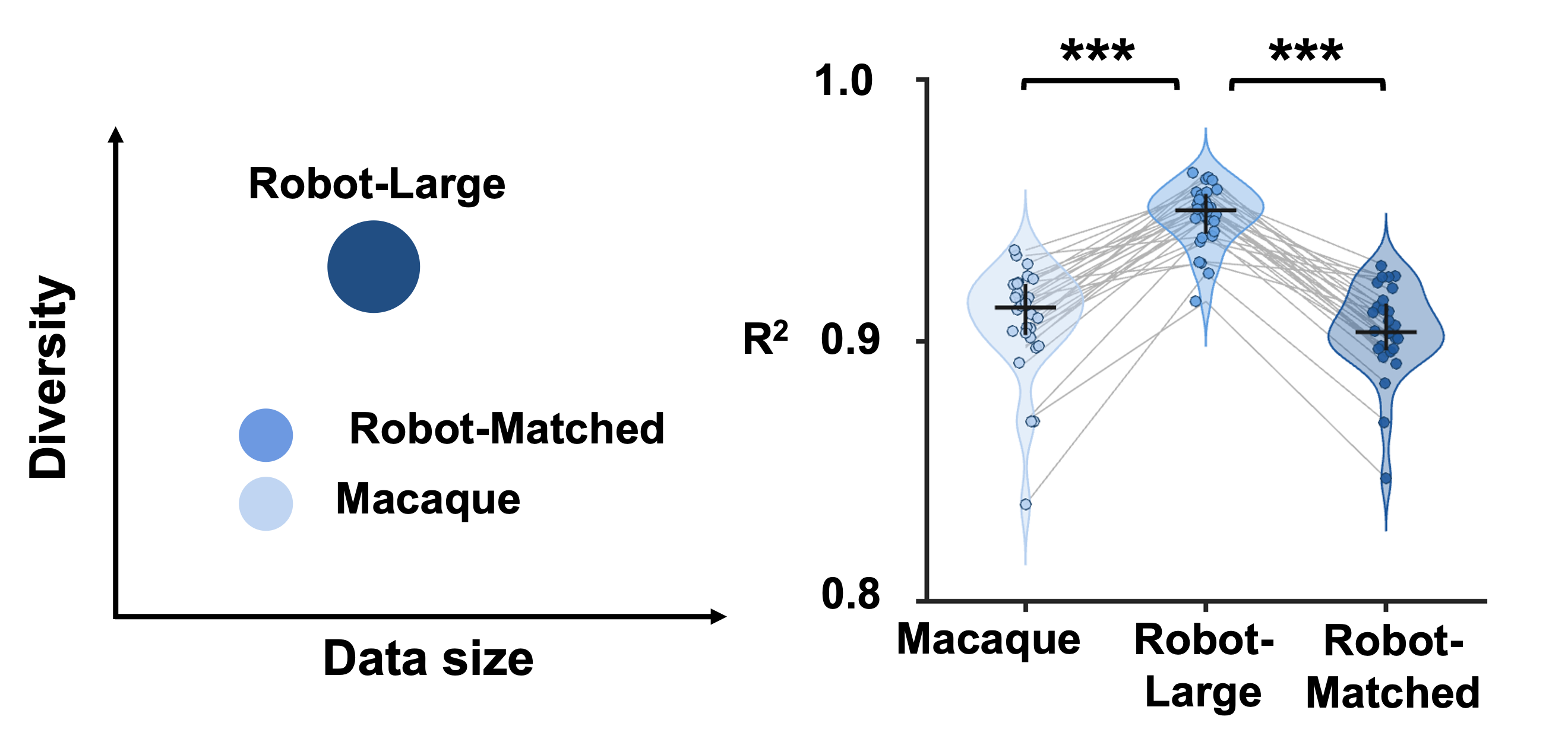}
\caption{The left panel schematically illustrates the relative pretraining-data scales. The right panel shows comparable performance for Macaque-PT and Robot-Matched-PT, while Robot-Large-PT achieves the best decoding performance.}
\label{fig4}
\end{figure}

\subsection{Scaled Pretrained Motor Encoder}
We next examined whether the improvements from behavioral pretraining were determined primarily by the source of the trajectories or by the scale of the pretraining data. We compared three pretrained motor-encoder initializations: Macaque-PT, Robot-Matched-PT, and Robot-Large-PT. Macaque-PT was pretrained on macaque trajectories, whereas Robot-Matched-PT was pretrained on trajectories from LIBERO-Spatial. The macaque and LIBERO-Spatial pretraining sets were matched in size, allowing the influence of behavioral source to be evaluated while approximately controlling for data scale. Robot-Large-PT was instead pretrained on the substantially larger LIBERO-100 dataset, enabling us to evaluate the effect of scaling robotic trajectory pretraining. Notably, the left panel of Fig. \ref{fig4} provides only a conceptual illustration of the relative pretraining conditions and dataset scales. It is not derived from the measured decoding scores.

As shown quantitatively in the right panel of Fig. \ref{fig4}, Macaque-PT and Robot-Matched-PT achieved comparable decoding performance despite the considerable difference between biological reaching movements and robotic manipulation trajectories. This result suggests that the motor encoder learns transferable kinematic structures that are not restricted to the system from which the trajectories were collected. More importantly, Robot-Large-PT provided the stronger overall performance, outperforming both scale-matched conditions. Thus, although macaque and robotic trajectories are both effective sources of behavioral supervision, increasing the scale and diversity of robotic pretraining data produces a stronger and more transferable behavioral representation. These findings support the premise of NeuroPB: readily available behavioral data from artificial systems can complement scarce neural recordings, and its benefit can be further increased by scaling and enriching behavioral pretraining.

\begin{figure}[t]
\centering
\includegraphics[width=1.0\columnwidth]{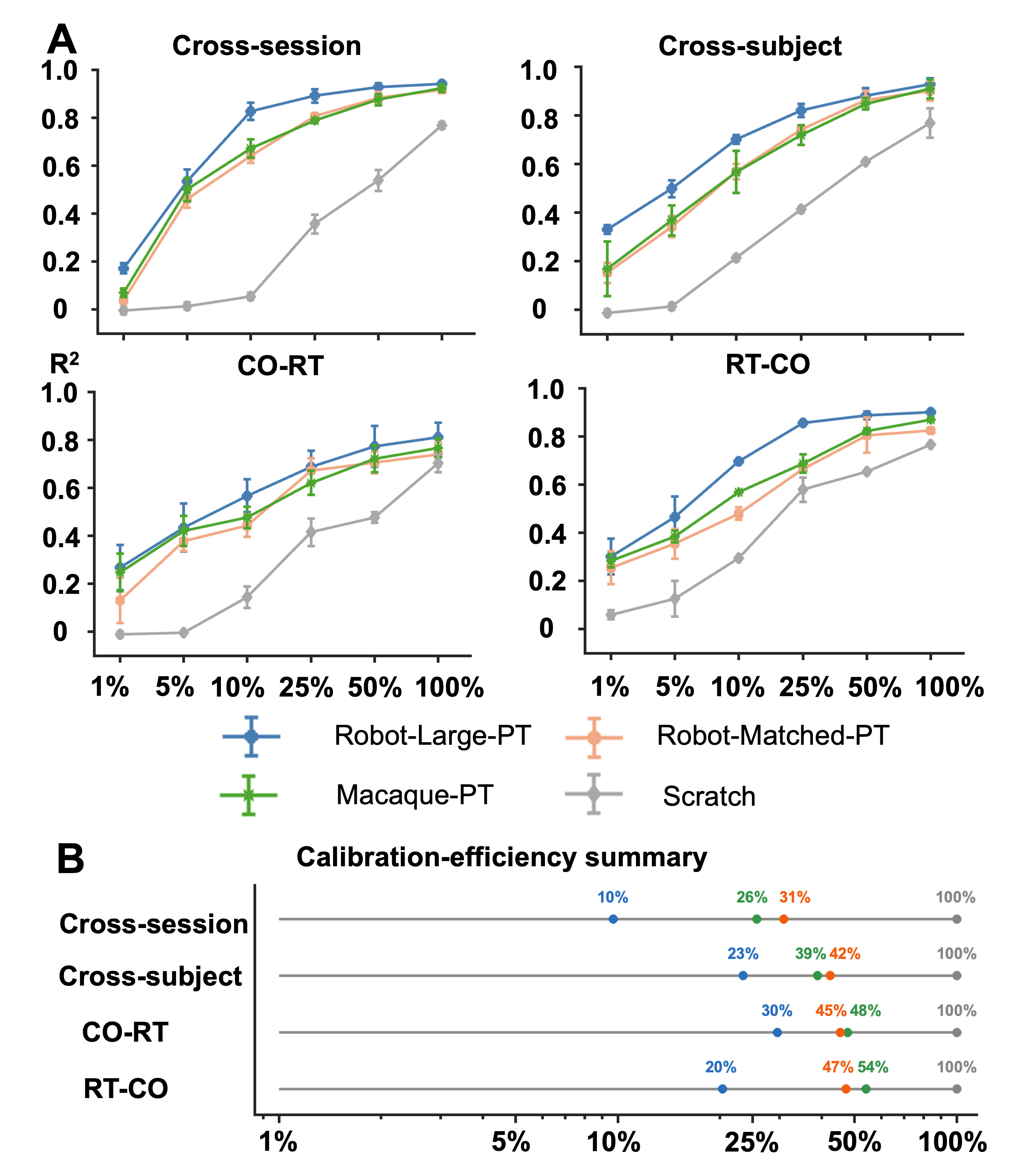}
\caption{(A) Calibration-efficiency validation under cross-session, cross-subject, cross-task including CO-RT and RT-CO. (B) Estimated calibration fraction required by each pretrained model to match the performance of Scratch using 100\% of the data.}
\label{fig5}
\end{figure}

\subsection{Calibration-Efficiency Validation}
We next evaluated whether behavioral pretraining reduces the amount of target-domain neural data required for adaptation across sessions, subjects, and motor tasks; the result is shown in Fig. \ref{fig5}A. Under each transfer setting, only the newly introduced unit and session embeddings were optimized using the fraction of target training data. In contrast, the scratch baseline initialized the entire model randomly and optimized all components using each calibration subset. More detailed calibration settings are provided in the Appendix. Across all four settings, the Robot-Large-PT achieved the strongest performance in the low-data regime. Its advantage was particularly evident with only 1\%–10\% of the target data, where it consistently exceeded the Robot-Matched-PT, Macaque-PT, and scratch models. The performance gap gradually narrowed as more calibration data became available, but Robot-Large-PT remained competitive at full calibration.

The calibration-efficiency summary in Fig. \ref{fig5}B further quantifies this advantage by estimating the fraction of target data required to match the performance of the scratch model trained on 100\% of the target data. Robot-Large-PT required less calibration data for generalization. In comparison, Macaque-PT and Robot-Matched-PT share a similar fraction of calibration data, and exhibited similar generalization and both substantially improved calibration efficiency relative to training from scratch. Increasing the scale and diversity of behavioral pretraining data produces more transferable behavioral representations, even when the pretraining trajectories originate from robotic rather than biological systems.

\begin{figure}[t]
\centering
\includegraphics[width=1.0\columnwidth]{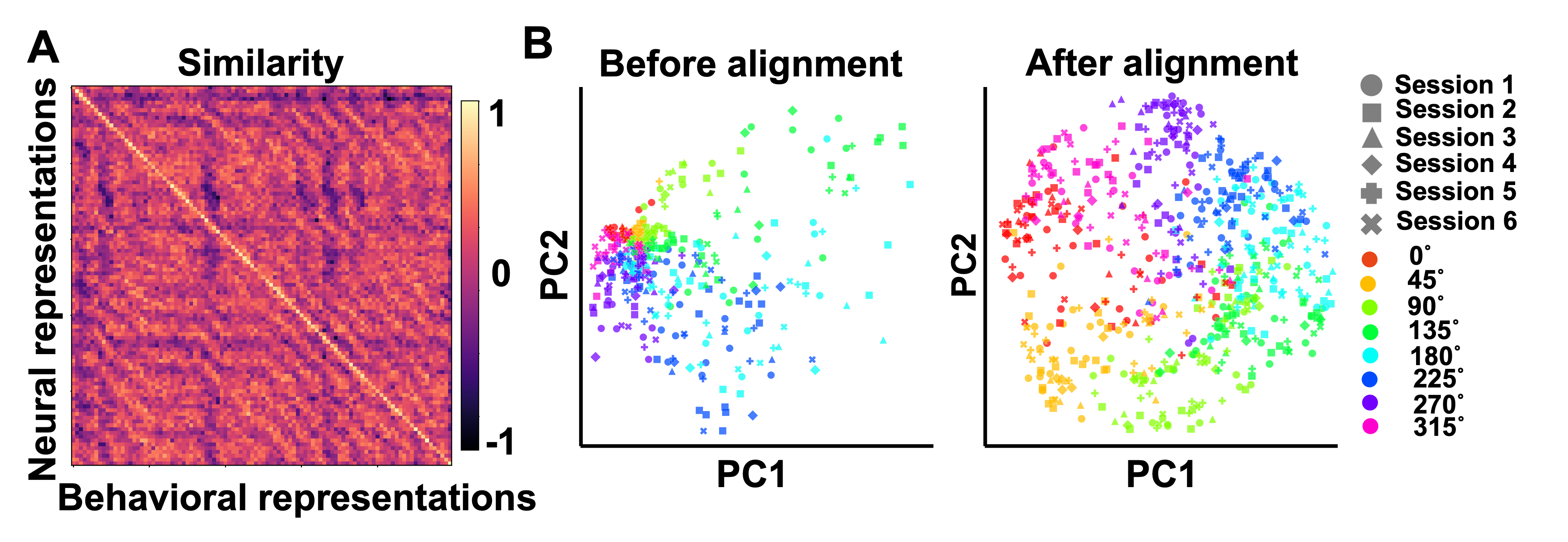}
\caption{(A) Pairwise similarity between neural and behavioral representations. (B) PCA visualization of neural representations before and after alignment, colored by movement direction and marked by recording session. Alignment improves directional organization while reducing session-dependent variation.}
\label{fig6}
\end{figure}

\subsection{Visualization Analysis}
To examine whether neural activity was aligned with the pretrained behavioral space, we computed the pairwise cosine similarity between neural and behavioral representations. The resulting similarity matrix in Fig. \ref{fig6}A exhibited a clear diagonal structure, indicating that each neural representation was most similar to its corresponding behavioral representation, while mismatched pairs generally showed lower similarity. This pattern provides direct representation-level evidence that the alignment objective successfully maps neural activity into a behaviorally meaningful embedding space.

To further examine how motor alignment reshapes the learned neural representation, we visualized held-out neural windows from six recording sessions using PCA in Fig. \ref{fig6}B. Before alignment, the neural representations were dominated by session-specific variability: samples from different recording sessions showed noticeable separation, whereas representations associated with different reach directions were substantially intermixed. After projection into the pretrained behavioral representation space, samples became more clearly structured according to reach direction, while marker shapes corresponding to different sessions were broadly intermingled within the same direction-related regions. These results suggest that motor alignment suppresses session-dependent nuisance variation while preserving behaviorally relevant movement information, thereby producing a more session-invariant neural representation.

\subsection{Ablation Studies}
We conducted ablation experiments on the CO dataset to quantify the contribution of contrastive alignment and the key components of the neural encoder, as summarized in Table \ref{table2}. We validated that removing the motor encoder structure produced the largest degradation, with $R^2$ of 0.8296 $\pm$ 0.0424. This substantial decrease, together with the increased performance variability, demonstrates that alignment with the pretrained behavioral representation space is critical for learning accurate and stable neural representations. Among the neural-encoder components, removing unit encoding caused the largest performance reduction, yielding an $R^2$ of 0.8825$\pm $0.0155, which highlights the importance of preserving neuron-specific information. Removing RoPE reduced performance to 0.9011$\pm $0.0090, confirming the value of explicitly modeling the relative temporal structure of spike sequences. Excluding latent self-attention resulted in an $R^2$ 
of 0.9065$\pm $0.0133, indicating that interactions among latent tokens contribute to population-level representation learning. Removing cross-attention produced a smaller but consistent decrease to 0.9182$\pm $0.0107, demonstrating its role in selectively aggregating spike-token information. Overall, these results show that behavioral contrastive alignment provides the largest contribution, while unit identity, temporal encoding, cross-attention, and latent self-attention jointly support effective neural trajectory decoding.
 

\begin{table}[t]
\centering
\begin{tabular}{l l l}
\toprule Methods & CO ($R^2$)  \\
\midrule w/o contrastive & 0.8296$\pm$0.0424\\
w/o unit encoding & 0.8825$\pm$0.0155\\
w/o RoPE & 0.9011$\pm$0.0090\\
w/o cross attention & 0.9182$\pm$0.0107\\
w/o self attention &0.9065$\pm$0.0133\\
Overall &0.9478$\pm$0.0120\\
\bottomrule
\end{tabular}
\caption{Ablation performance}
\label{table2}

\end{table}


\subsection{Conclusion}
We introduced NeuroPB, a neural decoding framework that transfers knowledge from pretrained behavioral representations to motor trajectory decoding. By aligning neural activity with a frozen behavioral representation space, NeuroPB consistently outperformed its scratch-trained counterpart. Macaque and scale-matched robotic trajectories provided comparable pretraining benefits, while large-scale robotic pretraining achieved the best performance, improved generalization, and calibration efficiency across sessions, subjects, and tasks. This study is limited to offline evaluation using invasive macaque recordings. Future work will extend behavioral pretraining and neural alignment to human recordings, including non-invasive neural signals, and evaluate NeuroPB in real-time closed-loop applications.




\bibliography{aaai2027}


\end{document}


\maketitle
\section{Experimental Details}
\subsection{Details of robotic pretraining datasets}
Fig. 4 provides the conceptual illustration of the relative pretraining conditions and dataset scales. It is not derived from the measured decoding scores. Here, I will explain the definition of scale and diversity. Robot-Matched-PT was pretrained using LIBERO-Spatial, which contains 10 language-conditioned manipulation tasks designed to vary the spatial relationships and layouts of a common set of objects. Each task includes 50 human-teleoperated demonstrations, yielding 500 robotic trajectories in total. This dataset was comparable in scale to the macaque trajectory set and therefore enabled the influence of behavioral source to be examined while approximately controlling for pretraining-data size. 

Robot-Large-PT was instead pretrained using LIBERO-100, which contains 100 manipulation tasks with 50 demonstrations per task, yielding 5,000 trajectories—ten times the scale of LIBERO-Spatial. LIBERO-100 additionally covers a substantially broader range of object configurations, spatial layouts, manipulation goals, and long-horizon behaviors. Each LIBERO demonstration provides synchronized workspace- and wrist-camera observations, proprioceptive states, robot actions, and a language task specification. After the preprocessing procedure of CoTracker, the demonstration videos were converted into point-track trajectories using an off-the-shelf visual tracker and subsequently used to pretrain the motor encoder. Thus, the comparison between Robot-Matched-PT and Macaque-PT primarily evaluates the effect of behavioral source, whereas the comparison between Robot-Matched-PT and Robot-Large-PT evaluates the combined benefit of increased robotic data scale and behavioral diversity.

\begin{algorithm}[t]
\caption{Training and inference of NeuroPB}
\label{alg:neuropb}
\begin{algorithmic}[1]
\REQUIRE Paired neural--trajectory data
$\{(\mathcal{S}_i,\mathbf{P}_i,s_i)\}_{i=1}^{B}$

\STATE \textbf{\textsc{Neural--Motor Alignment and Decoding}}
\FOR{each paired mini-batch}
    \FOR{$i=1,\ldots,B$}
        \STATE $\mathbf{x}_{i,j}
        \gets E_{\mathrm{unit}}(u_{i,j})
        +E_{\mathrm{session}}(s_i)$
        \STATE $\mathbf{Z}_{i,1}
        \gets \operatorname{CrossAttn}_{\mathrm{RoPE}}
        (\mathbf{Z}_{i,0},\mathbf{X}_i)$
        \STATE $\mathbf{Z}_{i,L}
        \gets \operatorname{SelfAttnBlocks}(\mathbf{Z}_{i,1})$
        \STATE $\widetilde{\mathbf{z}}_i^{n}
        \gets g_n(\operatorname{Pool}(\mathbf{Z}_{i,L}))$
        \STATE $\overline{\mathbf{z}}_i^{n}
        \gets \operatorname{Norm}(\widetilde{\mathbf{z}}_i^{n})$

        \STATE $\mathbf{H}_i^{m}
        \gets f_{\theta_m}(\mathbf{E}_i^{m})$
        \STATE $\overline{\mathbf{z}}_i^{m}
        \gets \operatorname{Norm}
        (\operatorname{Pool}(\mathbf{H}_i^{m}))$

        \STATE $\widehat{\mathbf{P}}_i
        \gets d_{\psi}(\overline{\mathbf{z}}_i^{n})$
    \ENDFOR

    \STATE $s_{ij}
    \gets (\overline{\mathbf{z}}_i^{n})^{\top}
    \overline{\mathbf{z}}_j^{m}/\tau$

    \STATE $\mathcal{L}_{\mathrm{align}}
    \gets \frac{1}{2}
    (\mathcal{L}_{n\rightarrow m}
    +\mathcal{L}_{m\rightarrow n})$

    \STATE $\mathcal{L}_{\mathrm{dec}}
    \gets \frac{1}{B}\sum_i
    \|\widehat{\mathbf{P}}_i-\mathbf{P}_i\|_2^2$

    \STATE $\mathcal{L}
    \gets \mathcal{L}_{\mathrm{dec}}
    +\mathcal{L}_{\mathrm{align}}$

    \STATE Update $\theta_n$, $g_n$, and $\psi$
\ENDFOR

\STATE
\STATE \textbf{\textsc{Inference}}
\STATE $\mathbf{Z}_{L}\gets f_{\theta_n}(\mathcal{S})$
\STATE $\overline{\mathbf{z}}^{n}
\gets g_n(\operatorname{Pool}(\mathbf{Z}_{L}))$
\STATE $\widehat{\mathbf{P}}
\gets d_{\psi}(\overline{\mathbf{z}}^{n})$
\STATE \textbf{return} $\widehat{\mathbf{P}}$

\end{algorithmic}
\end{algorithm}

\subsection{Motor Encoder Architecture}
The pretrained motor-trajectory encoder used an eight-layer Transformer with a hidden dimension of 384, eight attention heads, a patch size of 4, and attention and feed-forward dropout rates of 0.2. For each of the 100 trajectory time points, a local segment containing 16 consecutive two-dimensional coordinates was constructed. Each 16-step segment was divided into four non-overlapping patches, producing four 384-dimensional latent tokens. The motor encoder was kept frozen during alignment and motor-decoder training.

\subsection{Neural Encoder Architecture}
The neural encoder processed 1-s neural windows using a latent interval of 0.125~s and 16 latent tokens per interval, yielding \(N=128\) latent tokens per window. The encoder had a hidden dimension of \(D=64\) and contained \(L=6\) latent self-attention layers. It used eight self-attention heads, two cross-attention heads, and an attention-head dimension of 64. The feed-forward, linear, and attention dropout rates were 0.2, 0.4, and 0.2, respectively.

\subsection{Projection Layer Architecture}
Separate projection heads were used to map the neural and motor representations into a shared 3,200-dimensional space. On the neural side, the 128 latent-token representations were averaged to obtain a 64-dimensional representation, which was projected through a two-layer multilayer perceptron with dimensions 3200, a ReLU activation, and dropout of 0.3. On the motor side, the four 384-dimensional tokens associated with each local trajectory segment were averaged, producing one 384-dimensional representation at each of the 100 trajectory time points. A learnable projection layer mapped each representation from 384 to 32 dimensions. Concatenating the resulting 100 32-dimensional features produced a 3,200-dimensional motor representation. The neural and motor representations were normalized before computing the alignment losses. Finally, the motor decoder projected the 3,200-dimensional representation through a multi-layer perception to a 200-dimensional output, which was reshaped into 100 two-dimensional motor predictions.

\begin{figure}[!t]
\centering
\includegraphics[width=1.0\columnwidth]{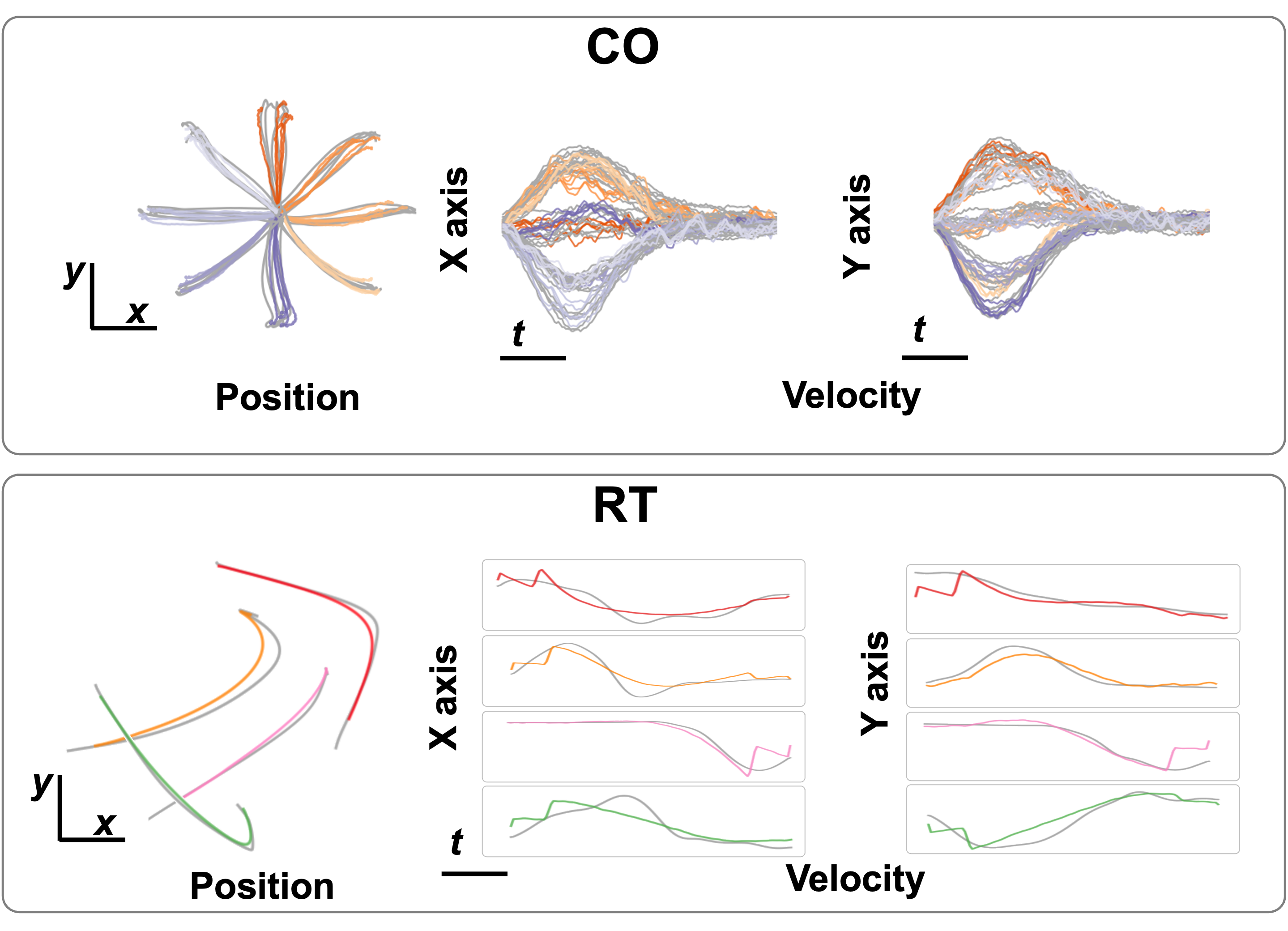}
\caption{Trajectory decoding results on the CO and RT tasks. The left panels show predicted and ground-truth two-dimensional position trajectories, while the middle and right panels show the corresponding x- and y-axis velocity profiles over time. The predictions closely follow the ground truth for both the structured center-out movements and the more complex random-target trajectories.}
\label{fig7}
\end{figure}

\subsection{Training and Optimization Details}
Training used a batch size of 128 and the SparseLamb optimizer with a weight decay of \(10^{-4}\). The learning rate was controlled by a one-cycle cosine schedule, with a warm-up fraction of 0.05 and a final learning-rate division factor of 100. According to the default configuration and implementation, the maximum learning rate was \(3.125\times10^{-4}\times128=0.04\). The contrastive temperature was learnable and initialized to \(\tau=0.07\). The objective combined the motor-decoding loss and a symmetric cross-entropy contrastive loss, with default weights of 1.

\subsection{Calibration Details}
In the transfer experiments, calibration was performed using only the labeled fraction of the training set from the target sessions. The pretrained representation encoder was initialized from the source-domain checkpoint and kept fixed, while the unit and session embeddings encoding in the neural encoder were optimized on the calibration data. Model selection and early stopping were based exclusively on the whole validation data from the target sessions, after which the calibrated model was evaluated once on the remaining held-out target-session test data. Thus, no target-test samples or labels were used for parameter fitting, hyperparameter selection, or checkpoint selection.

\begin{figure}[!t]
\centering
\includegraphics[width=1.0\columnwidth]{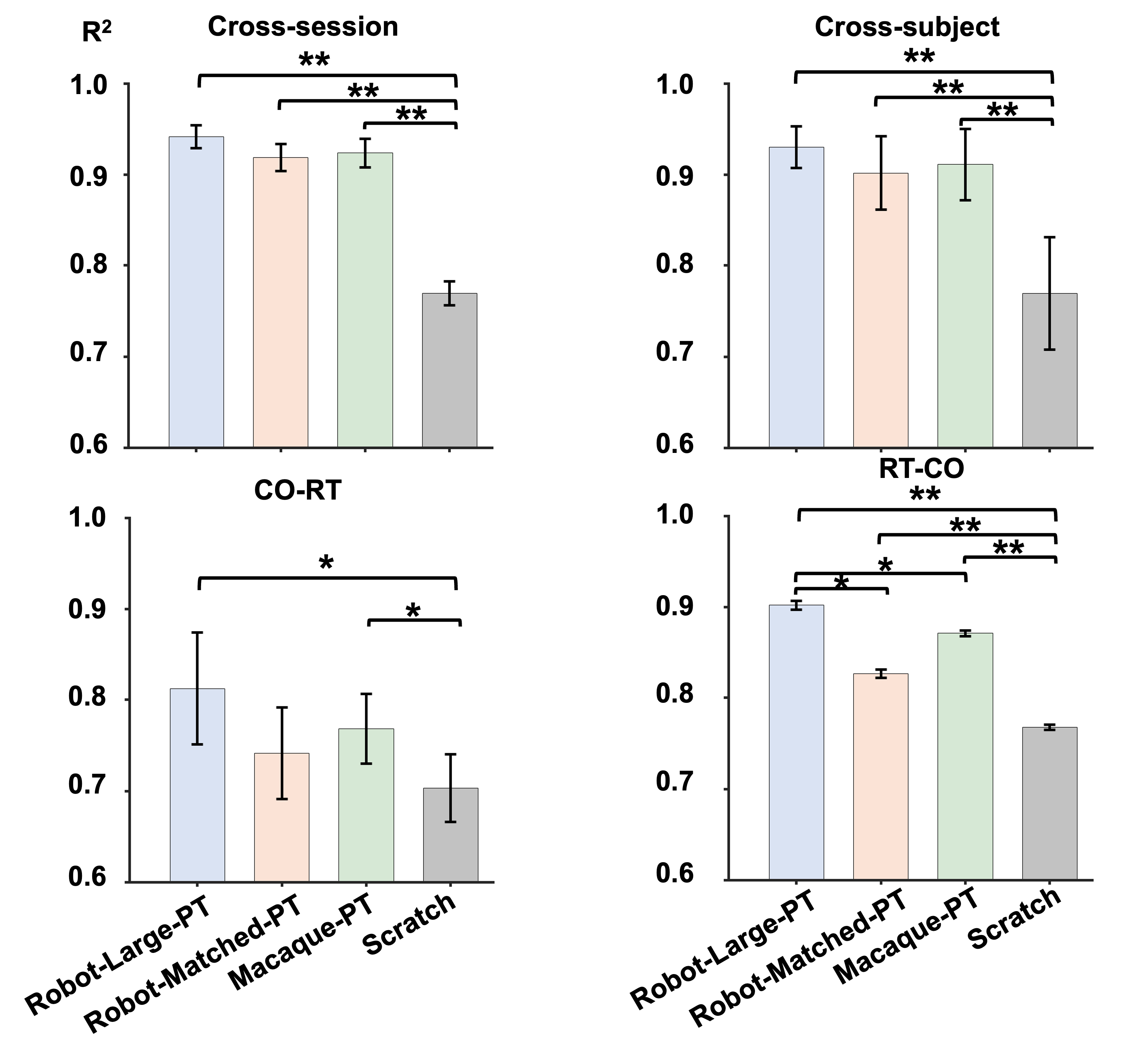}
\caption{Generalization performance is reported under cross-session, cross-subject, CO-RT, and RT-CO transfer settings. Bars represent the mean $R^2$ obtained using Robot-Large-PT, Robot-Matched-PT, Macaque-PT, and Scratch, and error bars indicate the standard deviation. Statistical comparisons are performed between each pretrained condition and Scratch. $^{*}p<0.05$ and $^{**}p<0.01$.}
\label{fig8}
\end{figure}

\section{Additional Results}
\subsection{Decoding Results}
Fig. \ref{fig7} shows more decoding results. It compares the trajectories predicted by NeuroPB with the corresponding ground-truth movements on the center-out (CO) and random-target (RT) tasks. For each task, the left panel shows the reconstructed two-dimensional position trajectories, while the middle and right panels present the temporal profiles of the velocity components along the x- and y-axes, respectively. The predicted trajectories closely follow the ground truth across both the regular, direction-specific movements in CO and the more complex, variable trajectories in RT, demonstrating the model’s ability to recover both spatial movement patterns and temporal dynamics.

\subsection{Generation Result}
Fig. \ref{fig8} reports the generalization performance, i.e., when 100\% of the available target-domain calibration data were used. In both the cross-session and cross-subject settings, all behaviorally pretrained models significantly outperformed the Scratch condition, demonstrating that behavioral pretraining remains beneficial even with full target-domain calibration. The improvement was also observed under cross-task transfer. For CO-RT transfer, Robot-Large-PT and Macaque-PT achieved significant improvements over Scratch, whereas Robot-Matched-PT showed a smaller, non-significant improvement. For RT-CO transfer, all pretrained conditions significantly outperformed Scratch. Moreover, Robot-Large-PT significantly outperformed both Robot-Matched-PT and Macaque-PT, indicating an additional benefit from larger scale and diversity of robotic pretraining. Overall, Robot-Large-PT achieved the strongest generalization performance across the evaluated settings, supporting the benefit of large-scale behavioral pretraining for generalization across recording sessions, subjects, and motor tasks.